\documentclass[conference]{IEEEtran}
\usepackage{times}
\pdfoutput=1 

\IEEEoverridecommandlockouts

\usepackage[svgnames]{xcolor}
\usepackage{verbatim}

\newcommand{\etal}{\emph{et al.}\xspace}
\newcommand{\etc}{\emph{etc.}\xspace}
\newcommand{\eg}{\emph{e.g.}\xspace}
\newcommand{\ie}{\emph{i.e.}\xspace}

\newcommand{\figLabel}{Fig.\xspace}

\newcommand{\secLabel}{Section\xspace}
\newcommand{\tblLabel}{Table\xspace}

\definecolor{orange}{rgb}{1,0.5,0}
\definecolor{maroon}{rgb}{0.51,0,0}

\usepackage{times}
\usepackage{epsfig}
\usepackage{graphicx}
\usepackage{amsmath}
\usepackage{amssymb}
\usepackage{xspace}
\usepackage[font=small]{caption}
\usepackage[linesnumbered,ruled,vlined]{algorithm2e}

\usepackage{subcaption}
\usepackage[font=small]{caption}
\usepackage{multirow}

\usepackage[labelformat=simple]{subcaption}

\usepackage[flushleft]{threeparttable}

\usepackage[utf8]{inputenc} 
\usepackage[T1]{fontenc}    
\usepackage{url}            
\usepackage{booktabs}       
\usepackage{amsfonts}       
\usepackage{nicefrac}       
\usepackage{microtype}      
\usepackage{color}
\usepackage{mathtools}
\usepackage{bbm}
\usepackage{bm}
\usepackage{braket}
\usepackage{xspace}

\usepackage{subcaption} 
\usepackage{caption}
\usepackage{wrapfig}

\usepackage{amsthm,commath}
\usepackage[utf8]{inputenc}
\usepackage[english]{babel}

\usepackage{amsmath,amsfonts,bm}

\def\eqref#1{equation~\ref{#1}}

\def\1{\bm{1}}

\def\ve{{\bm{e}}}
\def\vf{{\bm{f}}}

\def\vr{{\bm{r}}}

\def\vv{{\bm{v}}}
\def\vw{{\bm{w}}}
\def\vx{{\bm{x}}}

\def\mD{{\bm{D}}}
\def\mE{{\bm{E}}}
\def\mF{{\bm{F}}}

\def\mI{{\bm{I}}}

\def\mM{{\bm{M}}}

\def\mP{{\bm{P}}}

\def\mV{{\bm{V}}}

\DeclareMathAlphabet{\mathsfit}{\encodingdefault}{\sfdefault}{m}{sl}
\SetMathAlphabet{\mathsfit}{bold}{\encodingdefault}{\sfdefault}{bx}{n}

\def\gG{{\mathcal{G}}}

\theoremstyle{plain}

\theoremstyle{definition}

\theoremstyle{remark}

\usepackage{xspace}
\usepackage{comment}
\usepackage{balance}
\usepackage{etoolbox,siunitx}
\usepackage{calc}
\usepackage{pifont,hologo}

\usepackage[super]{nth}
\usepackage{nicefrac}
\def\tt{\mathbf{t}}

\def\vv{\mathbf{v}}

\def\gG{\mathcal{G}}

\def\mM{\mathcal{M}}

\def\Re{\mathbb{R}}

\DeclareMathSymbol{@}{\mathord}{letters}{"3B}

\def\latex/{\LaTeX}
\def\bibtex/{\hologo{BibTeX}}

\usepackage[official]{eurosym}

\SetKwComment{Comment}{/* }{ */}

\newcommand{\GL}[1]{{\bf \color{green}[GL: #1]}}

\newcommand{\JC}[1]{{\bf \color{blue}[JC: #1]}}

\usepackage[hmargin=2cm,vmargin=2.5cm]{geometry}
\usepackage{etoolbox}
\usepackage{hhline}
\usepackage{tablefootnote}
\usepackage{pdflscape}
\usepackage{colortbl}%
\newcommand{\myrowcolor}{\rowcolor[gray]{0.925}}
\usepackage{booktabs}

\newif\ifblackandwhite
\ifblackandwhite
  
  \newcommand{\highest}[1]{\textbf{#1}}
\else

  \newcommand{\highest}[1]{\textcolor{Maroon}{\textbf{#1}}}%
\fi

\usepackage[frozencache=true,cachedir=.]{minted} 

\definecolor{bg}{rgb}{0.95,0.95,0.95}

\usepackage[numbers]{natbib}
\usepackage{multicol}
\usepackage[bookmarks=true]{hyperref}

\title{\LARGE \bf
How To Not Train Your Dragon: Training-free Embodied Object Goal Navigation with Semantic Frontiers
}

\author{Author Names Omitted for Anonymous Review. Paper-ID [143]}
\author{
    Junting Chen$^{1*}$, Guohao Li$^{2*}$, Suryansh Kumar$^{1\dagger}$, Bernard Ghanem$^{2}$, Fisher Yu$^{1}$
    \thanks{$^{*}$First two authors share equal contributions.}
    \thanks{$^{1}$ ETH Z\"urich}, \thanks{$^{2}$ King Abdullah University of Science and Technology (KAUST)}
    \thanks{$^{\dagger}$ Corresponding Author (k.sur46@gmail.com)}
}
\begin{document}
\maketitle
\thispagestyle{empty}
\pagestyle{empty}

\newcommand{\kumar}[1]{\textcolor{red}{\textbf{Kumar:} #1}}

\begin{abstract}
Object goal navigation is an important problem in Embodied AI that involves guiding the agent to navigate to an instance of the object category in an unknown environment---typically an indoor scene. Unfortunately, current state-of-the-art methods for this problem rely heavily on data-driven approaches, \eg, end-to-end reinforcement learning, imitation learning, and others. Moreover, such methods are typically costly to train and difficult to debug, leading to a lack of transferability and explainability. Inspired by recent successes in combining classical and learning methods, we present a modular and training-free solution, which embraces more classic approaches, to tackle the object goal navigation problem. Our method builds a structured scene representation based on the classic visual simultaneous localization and mapping (V-SLAM) framework. We then inject semantics into geometric-based frontier exploration to reason about promising areas to search for a goal object. Our structured scene representation comprises a 2D occupancy map, semantic point cloud, and spatial scene graph.
Our method propagates semantics on the scene graphs based on language priors and scene statistics to introduce semantic knowledge to the geometric frontiers. With injected semantic priors, the agent can reason about the most promising frontier to explore.  The proposed pipeline shows strong experimental performance for object goal navigation on the Gibson benchmark dataset, outperforming the previous state-of-the-art. We also perform comprehensive ablation studies to identify the current bottleneck in the object navigation task.
\end{abstract}

\section{Introduction}

In recent years, the focus on computer vision has moved from using ``internet data'' such as images, videos, texts, etc., towards developing an active vision system involving a robot or an agent that can perceive the 3D scene and act intelligently. Accordingly, current research trends in this direction have begun to advocate for building Artificial Intelligence (AI) that involves embodiment \cite{kolve2017ai2,Xia2018GibsonER,gan2021threedworld,Savva2019HabitatAP}. Among several popular tasks in Embodied-AI, object goal navigation (ObjectNav) is one of the most important and sought-after tasks to solve \cite{batra2020objectnav}. ObjectNav requires an agent to search for any specific object category instance in an unknown environment. Unlike classical visual navigation \cite{survey_visual_navigation}, this task setup does not provide the target's position. As a result, it requires the agent to understand the scene's geometry and other higher-level semantics \cite{chaplot2020object}.

\begin{figure*}[t]
    \centering
    \includegraphics[width=0.9\linewidth]{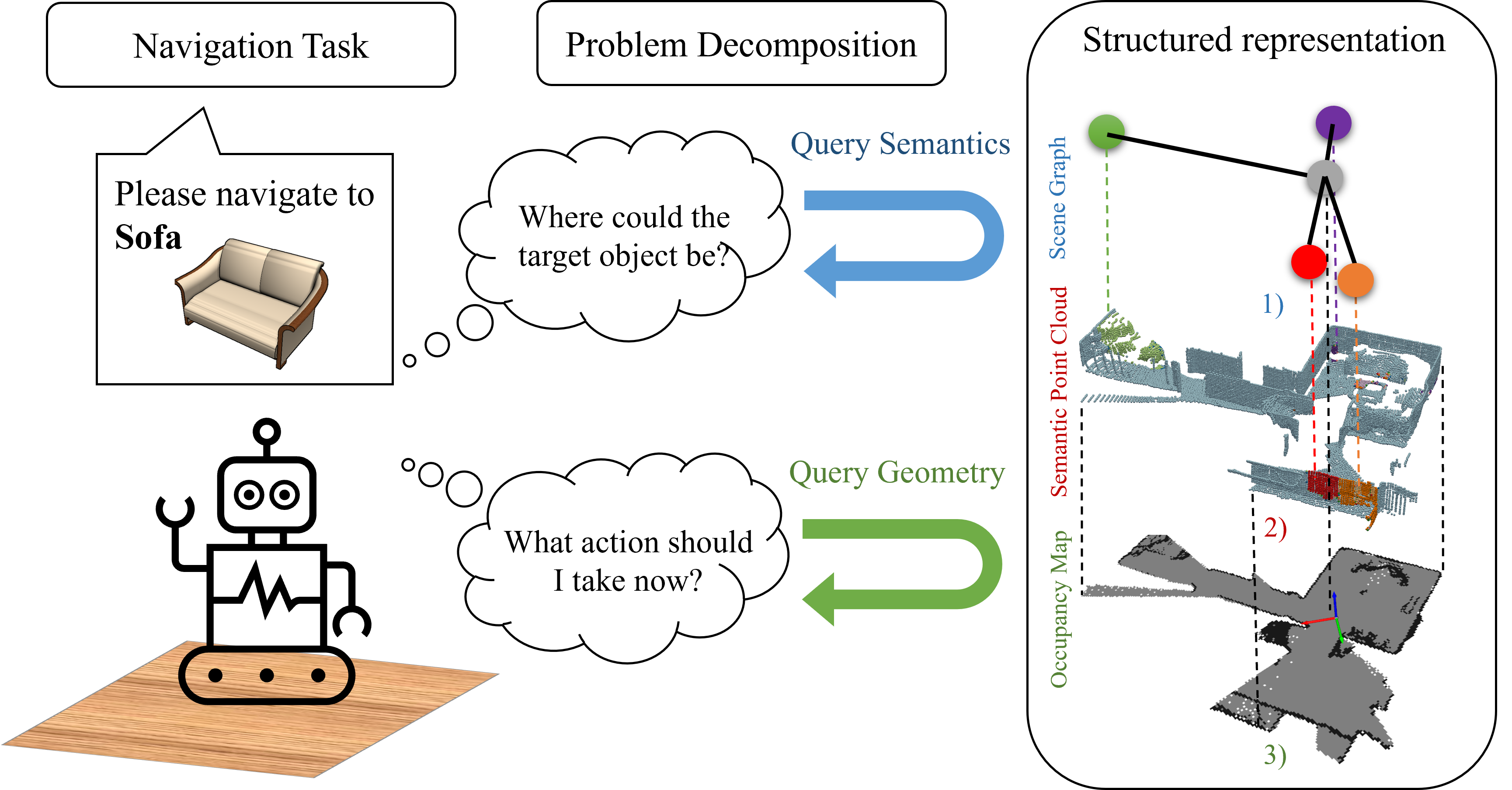}
    \caption{\textbf{Object Navigation with Structured Scene Representation.} ObjectNav can be decomposed into inferring the potential position of the target object in the scene and point-to-point planning. Provided a structured representation of the scene, which is composed of \textit{(i)} a spatial scene graph, \textit{(ii)} semantic point clouds, and \textit{(iii)} a 2D occupancy map, an agent can handle the two sub-tasks by querying semantic and geometric information from the scene graph and occupancy map separately. For clarity, the scene graph and occupancy map are computed from the semantic points cloud, thus the semantic point cloud is also considered part of our structured scene representation.
    }
    \label{fig.1}
\end{figure*}


To perform ObjectNav, one possible sequence of steps similar to human solutions is:  First, an agent must understand the target object such as its appearance, shape, etc. Next, the agent should explore the unseen environment while at the same time determining whether the target object is observed. If so, the agent takes the shortest path possible to reach the object while avoiding collisions. If not, it must resort to the favorable unexplored areas to unveil based on current information, \ie, knowing which parts of the scene are explored and reason about the likelihood of the object based on observations. Overall, an agent needs to have capabilities such as episodic memory to remember the unexplored parts of the scene, higher-level semantic priors to reason the next exploration location,  path planning to go to the object, and collision avoidance.

In the same vein, previous works attempt to solve this problem in two different ways. The \emph{first line of work} formulates it as an end-to-end learning problem. They try to learn both robot perception and control using reinforcement or imitation learning approaches.
Although end-to-end learning methods have shown some promising results on a few datasets, they typically have low sample efficiency, questionable generalization, and lack explainability; hence, it is hard to reason about its failure or success case and deploy them on real robots for practical use applications. The \emph{second line of work} combines classic navigation approaches with learning-based methods \cite{chaplot2020learning,chaplot2020object}. For instance, SemExp~\cite{chaplot2020object} uses a learning-based semantic mapping module based on Active Neural SLAM \cite{chaplot2020learning} to build a semantic grid map. Furthermore, a goal-oriented semantic policy is trained to predict long-term goals based on the semantic map using reinforcement learning. Fast Marching Method \cite{Janson2013FastMT}, an analytical path planner, is used to plan a path for the long-term goal. Finally, discrete actions are generated by a deterministic controller along the path. Compared to end-to-end methods, SemExp has better sample efficiency, generalization, and transferability to real-world scenarios. Yet, SemExp requires 10 million frames to train their network.



To mitigate the limitations of the current approaches, this work proposes a modular and training-free pipeline \textbf{StructNav}, which navigates an agent with a structured representation of the environment for the target object. Our method improves and enhances the current state of employing classic and learning-based modular approaches to solve this problem. Compared to SemExp \cite{chaplot2020object}, StructNav introduces a classic semantic visual SLAM to obtain a semantic point cloud map instead of a learned 2D semantic mapping module. Moreover, StructNav leverages a structured scene representation in which a 2D occupancy map for planning and a spatial scene graph for reasoning is constructed from the point cloud map. To avoid training an exploration policy with reinforcement learning, we propose a semantic frontier (SemFrontier) module, which combines classic frontier-based exploration with semantic priors. Our structured scene representation allows easy generation of proposal frontiers and propagating semantics to the frontiers. Specifically, we introduce language and scene-based priors to reason the promising unexplored areas by scoring geometric-based frontiers with semantics via the spatial scene graph. The language priors are acquired from pre-trained large-scale language models that encode knowledge from large-scale natural language inference datasets. The scene priors are obtained from the training split of 25 Gibson scenes, following the Batra et al. \cite{batra2020objectnav} experiment setting. 

From the perspective of real-world robotic systems setup, there is still an interesting sim-to-real gap for recent approaches in the object goal navigation domain, many of which adopt the ideal problem setting defined in \cite{Anderson2018OnEvalEmbodiedAI,batra2020objectnav}, where ground truth localization is provided. However, as a fundamental building block of robotic systems, localization significantly impacts downstream applications, including mapping and path planning. In this work, we also want to understand how this gap affects general visual navigation performance and how noise in other building blocks of the robotic pipeline affects performance. Thus, we propose a comprehensive benchmark for our modular method in which the ground truth output for each of our building blocks is turned on/off, including robot location, to assess how noise in the different building blocks of a robotic system affects the task of object goal navigation.

To summarize our key contributions are:
\begin{itemize}
    \item We introduce a modularized pipeline that reaches state-of-the-art performance for Object Goal Navigation, with a simple learning-free policy module, easily deployable to ROS-based robots.
    \item Our work propose a more realistic and comprehensive benchmark for visual navigation and discuss the impact of common building blocks of a robotic system on the performance of object goal navigation.  
\end{itemize}

\section{Related Work}
Our proposed method contributes to several modules in the overall pipeline for object goal navigation, including different perception and control aspects such as scene representation, navigation, \etc. Therefore, we discuss the works closely related to our proposed pipeline for brevity.






\noindent
\textbf{\textit{(a)} Classical Vision-Based Navigation.} These methods rely on the explicit map of the scene and generally perform point-to-point navigation. The agent explores the scene using frontier-based exploration algorithm \cite{yamauchi97frontier} and navigates along the computed optimal path to the goal using the well-crafted path planning algorithm \cite{LaValle1998RRT,sethian1996fast}. 

\smallskip
\noindent
\textbf{\textit{(b)} Object Goal Navigation.}
The end-to-end reinforcement learning (RL) based method has gained popularity for solving object goal navigation tasks in recent years \cite{savva2017minos, batra2020objectnav, Anderson2018OnEvalEmbodiedAI, wahid2021learning}. Current attempts have tried to improve the performance of such methods generally by using data-augmentation \cite{maksymets2021thda}, and higher-level scene representation \cite{druon2020visual,pal2021learning,du2020learning}. By higher-level scene representation, we mean object-level relation graph \cite{zhang2021hierarchical}, spatial attention map \cite{mayo2021visual}, semantic segmentation map \cite{mousavian2019visual}, and others \cite{zhu2021soon}. Lately, a combination of RL-based learning with classical methods has emerged, showing excellent performance accuracy compared to end-to-end learning methods \cite{ramakrishnan2022poni,chaplot2020object,liang2021sscnav}. Nevertheless, as mentioned before, popular methods along this line of approaches rely on RL-based semantic exploration module, which requires extensive training data. On the contrary, \cite{ramakrishnan2022poni} proposed a convolutional encoder-decoder module which is trained on 3D semantic segmentation dataset \cite{Xia2018GibsonER,chang2017matterport3d} to overcome the computational overhead with \cite{chaplot2020object,liang2021sscnav}.





 


\section{Methodology}

\secLabel \ref{sec:problem definition} provides a formal definition of the ObjectNav problem. In \secLabel \ref{sec:pipeline} an overview of the pipeline is outlined, specifically focusing on how data are processed through the pipeline. \secLabel \ref{sec:struct repsent}, contains details about how the structured representation is constructed. 
Lastly, \secLabel \ref{sec:navigation} describes how navigation goals and actions are calculated based on our structured scene representation.

\subsection{Problem Definition of Object Navigation} 
\label{sec:problem definition}



\begin{figure*}[ht]
    \centering
    \includegraphics[width=1\textwidth]{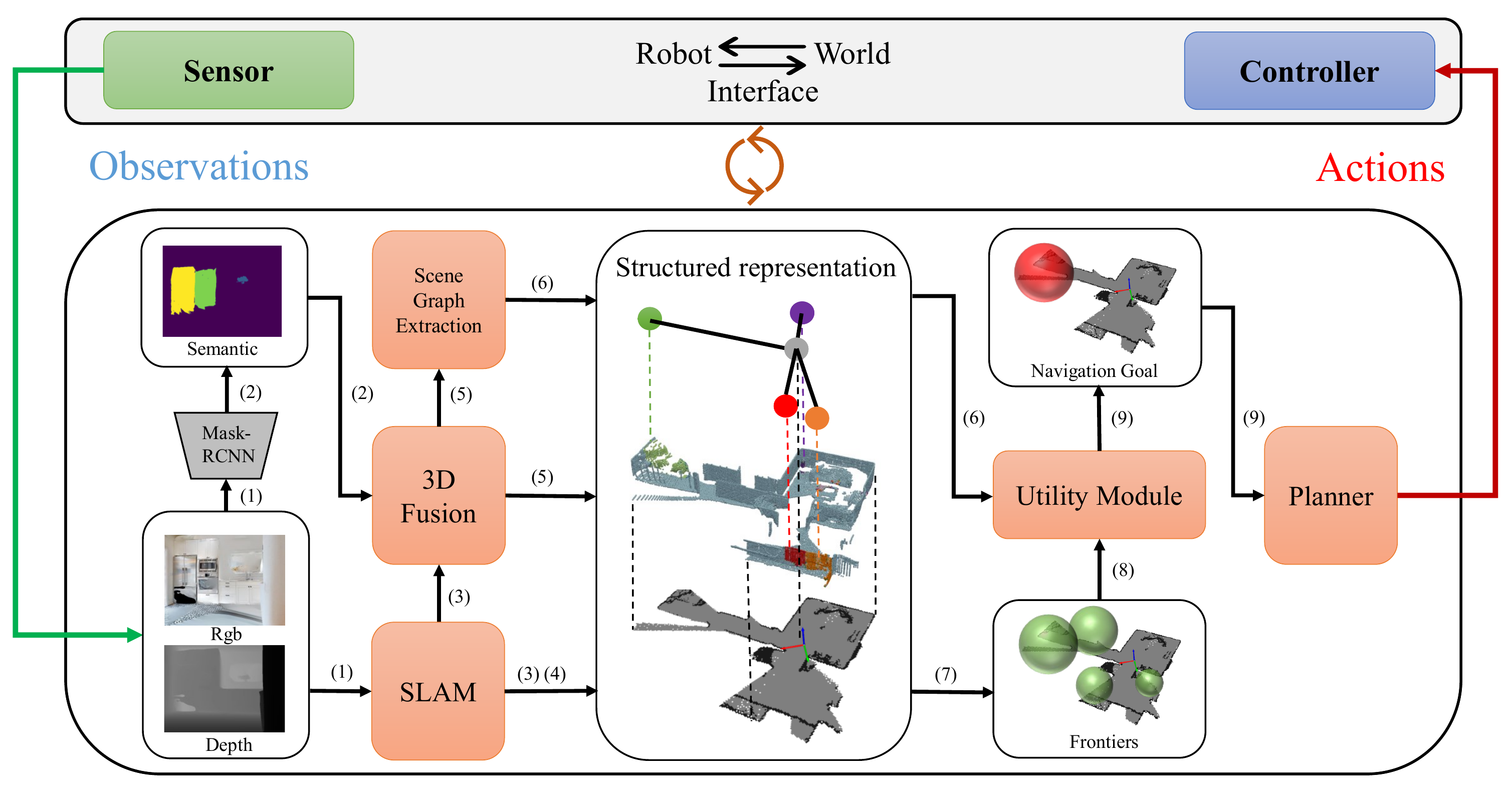}
    \caption{\textbf{Overview of the StructNav pipeline.} Our pipeline runs in the loop of receiving observations and generating actions to navigate an agent to the goal object in an unknown scene. Colored boxes shows functional modules and arrows represent data flows: (1) RGBD observation $\mI_t=(\mI_t^{\text{rgb}}$, $\mI_t^{\text{depth}})$ (2) Semantic Image $\mI_t^{sem}$ (3) Estimated pose $\hat{\bm{s}}_t^l$ (4) RGB point cloud $\mP_t^{\text{RGB}}$ (5) Semantic point cloud $\mP_t^{\text{sem}}$ (6) Spatial scene graph $\gG_t$ (7) 2D occupancy map $\mM_t$ (8) Frontiers $\mF_t$ (9) Intermediate navigation goal $\vx_t^{\text{goal}}.$
}\label{fig.2}
\end{figure*}

We formally define ObjectNav following \cite{batra2020objectnav}. An agent is randomly initialized on the floor in an unknown environment $\mathcal{E}$ with the initial pose $\bm{s}_0^{w}=( \vx_0^w, \vr_0^w)$, where $\vx_0^w\in\Re^3$ and 
$\vr_0^w\in SO(3)$
represent the initial position and the initial rotation in the world frame $w$, respectively. The agent is required to navigate to an instance of the goal object category $g$ specified by the category name (such as "chair").
We use the Habitat simulator \cite{Savva2019HabitatAP} as our testbed, in which the action space $\mathcal{A}$ is discretized into four actions: \texttt{turn\_left}, \texttt{turn\_right}, \texttt{move\_forward}, and \texttt{stop}. At time step $t$, the agent executes action $a_t$ and receives visual observation $\mI_t=(\mI_t^{\text{rgb}}, \mI_t^{\text{depth}})$ from a noiseless RGBD camera, where $\mI_t^{\text{rgb}} \in\Re^{H\times W\times 3}$ for the RGB image and $\mI_t^{\text{depth}}\in\Re^{H\times W}$ for the depth image.

In the experimental setting of SemExp \cite{chaplot2020learning}, the agent is equipped with a noiseless GPS+Compass sensor that provides the true relative pose information $\bm{s}_t^{l*}=(\vx_t^{l*}, \vr_t^{l*})$ to the initial state in the local frame $l$ at time $t$. In this paper, we also propose to evaluate ObjectNav in a more realistic setting, in which the localization from visual odometry on the fly replaces ground-truth localization. In this setting, the pose of an agent is annotated $\hat{\bm{s}_t^l}=(\hat{\vx_t^l}, \hat{\vr_t^l})$. 


\subsection{Structured Navigation Pipeline}
\label{sec:pipeline}
We propose a modular and \textit{training-free} pipeline named \textit{StructNav} to tackle the ObjectNav problem. StructNav uses a structured scene representation that consists of a semantic point cloud, a 2D occupancy map, and a spatial scene graph. With this structured representation, the semantic information for target position inference and the geometric information for planning on the 2D map can be decomposed and queried by different modules separately.


\figLabel \ref{fig.2} depicts how our training-free pipeline interacts with the 3D physical environment via receiving observations from sensors and executing planned actions. 
At each time step $t$, data flow starts at observations from the agent sensor and ends at actions sent to the agent controller. The data flow consists of two stages: perceiving to generate a structured scene representation and exploiting the structured representation to perform navigation actions.

In the first stage, our pipeline updates the structured representation of the scene by the current observation: 1) Given the visual observation $\mI_t$, the semantic segmentation image $\mI_t^{\text{sem}}$ is predicted from the input RGB image $\mI_t^{\text{rgb}}$ by a pre-trained Mask R-CNN \cite{he2017mask}; 2) The visual SLAM module receives the latest visual observation $\mI_t$, predicts the current agent pose $\hat{\bm{s}}_t^l$, and updates the dense RGB reconstruction of the scene $\mP_t^{\text{rgb}}$; 3) The dense RGB reconstruction of the scene is projected onto $xy-plane$ to generate a 2D occupancy map $\mM_t$; 4) Taken into the agent pose $\hat{\bm{s}}_t^l$, depth image $\mI_t^{\text{depth}}$, and the semantic image $\mI_t^{\text{sem}}$, the semantic point cloud $\mP_t^{\text{sem}}$ is updated by the 3D fusion module via back projection; 5) Spatial scene graph $\gG_t=(\mV_t, \mE_t)$ is extracted from the semantic point cloud $\mP_t^{\text{sem}}$. 

In the second stage, our pipeline computes an intermediate navigation goal in the map and generates an action to execute: 6) Frontiers $\mF_t=\{\vf_t^0, \cdots, \vf_t^{N_t}\}, \vf_t\in\Re^2$ are cluster centers of the boundary pixels between the explored area and the unexplored area on the 2D occupancy map, as Yamauchi \etal \cite{yamauchi97frontier} do; 7) Utility module combines the geometric information from the frontiers $\mF_t$ and semantic information from the spatial scene graph $\gG_t$ to select the most promising frontier as the navigation point goal $\vx_t^{\text{goal}}\in\mathbb{R}^{2}$. 8) The global planner estimate the path from the current agent pose $\hat{\bm{s}}_t^l$ to the point goal $\vx_t^{\text{goal}}$ and the local planner selects the correct action $a_t$ to follow the path at time $t$. 

For clarity, we provide our algorithmic approach in Pseudo Code \ref{pseudo-code}, demonstrating how data is processed in the pipeline. 



\begin{listing}[ht]
    \begin{minted}[tabsize=2, fontsize=\footnotesize, bgcolor=bg]{python}
def struct_nav(Env, SemSeg, SLAM, Fusion3D, 
  Planner, target):

  Finished = 0
  while not Finished:
    I_rgb, I_depth = Env.get_observation() 
    # first stage, update struct. representation
    I_sem, map, sg, pose = update_struct_repr(
    SemSeg, SLAM, Fusion3D, I_rgb, I_depth,target)
    # second stage, do navigation   
    nav_goal, Finished = navigate(I_depth, I_sem, 
      target, map, sg, pose)
    if not Finished:
      action = Planner.navigate(
        map, pose, nav_goal)
    else:
      action = "STOP"
    Env.execute_action(action)
  return 

def update_struct_repr(SemSeg, SLAM, Fusion3D, 
  I_rgb, I_depth):

  # process geometric data
  pcl_rgb, pose = SLAM.update(I_rgb, I_depth)
  map = SLAM.project_to_map(P_rgb) 
  frontiers = frontier_detect(map) 
  # process semantic data
  I_sem = SemSeg.predict(I_rgb)
  pcl_sem = Fusion3D.update(I_sem, pose) 
  sg = build_scene_graph(pcl_sem)
  return I_sem, map, sg, pose

def navigate(I_depth, I_sem, 
  target, map, sg, pose):

  Finished = 0
  if target in I_sem:
    # if target in this frame, simply moves to it
    nav_goal=get_object_goal(I_depth, I_sem, pose)
    Finished=check_goal_reached(pose, nav_goal)
  else:
    # explore the scene to look for the target 
    frontiers=get_frontiers(map)
    nav_goal=select_frontier(frontiers, sg, pose)
  return nav_goal, Finished
  
    \end{minted}
\caption{\textbf{StructNav}  (Python script). After receiving the RGBD images from camera sensors, {StructNav} first updates the structured representation by processing the geometric and semantic information. Then, {StructNav} enters the \emph{navigation} stage. The agent will move to the goal if the target is in this frame. Otherwise, the agent will navigate to the most promising frontier obtained from our structured representation.}
\label{pseudo-code}
\end{listing}

\subsection{Structured Scene Representation}
\label{sec:struct repsent}
As discussed in section \ref{sec:pipeline}, our proposed structured scene representation has the following components: 1) 2D occupancy map $\mM_t\in \Re^{h_t\times w_t}$, where $h_t$ and $w_t$ are the height and width of the occupancy map, which is a slack bound of the explored area, automatically expanded by the SLAM module; 2) semantic point cloud $\mP_t^{\text{sem}}\in \Re^{k_t\times4}$, where $k_t$ is the number of points in the point cloud at time $t$. Each point has four channels. The first three channels are point coordinates, and the last channel is the semantic label; 3) spatial scene graph $\gG_t=(\mV_t, \mE_t)$. $\mV_t=\{\vv_t^i\}$ are object nodes in the graph, and each node $\vv_t^i\in\Re^4$ has the same 4 channels as the semantic point cloud $\mP_t^{\text{sem}}$, representing the center of the object and its label. Edges $\mE_t=(\ve_t^{ij})$ are translations between object nodes, $\ve_t^{ij}\in \mathbb{R}^3$ represents the translation from $\vv_t^i$ to $\vv_t^j$. To construct this structured representation, we build our pipeline on top of a popular visual SLAM system RTAB-Map \cite{Labbe2019RTABMap}, which takes per-frame RGBD observations to predict the agent pose and construct the semantic point cloud on the fly. 
We use a purely geometry-based algorithm DBSCAN \cite{ester1996dbscan} to cluster the semantic point cloud, which clusters points based on the point density in the neighborhood. It is worth mentioning that we take the semantic label weighted by a large factor as the fourth dimension in DBSCAN to avoid points of different categories being classified into one cluster. The semantic segmentation model could produce false predictions and inter-frame inconsistency in 2d semantic images, leading to faulty object nodes in 3D. To reduce the false predictions, we compute the bounding boxes of all clusters and then use 3D non-maximum suppression (3DNMS) to filter out small clusters which share a high IOU rate with other large ones.


\subsection{Exploration with Semantic Frontiers}


With the structured scene representation, an agent explores the unknown environment with semantic reasoning on the spatial scene graph for target inference and path planning on the 2D grid map until the target is detected in the current observations. Then it simply navigates to the target object with the planner. In the rest of this section, we will focus on frontier-based exploration with semantic utilities. 

\begin{figure*}[h]
    \centering

     \begin{subfigure}[b]{0.48\textwidth}
         \centering
         \includegraphics[width=\textwidth]{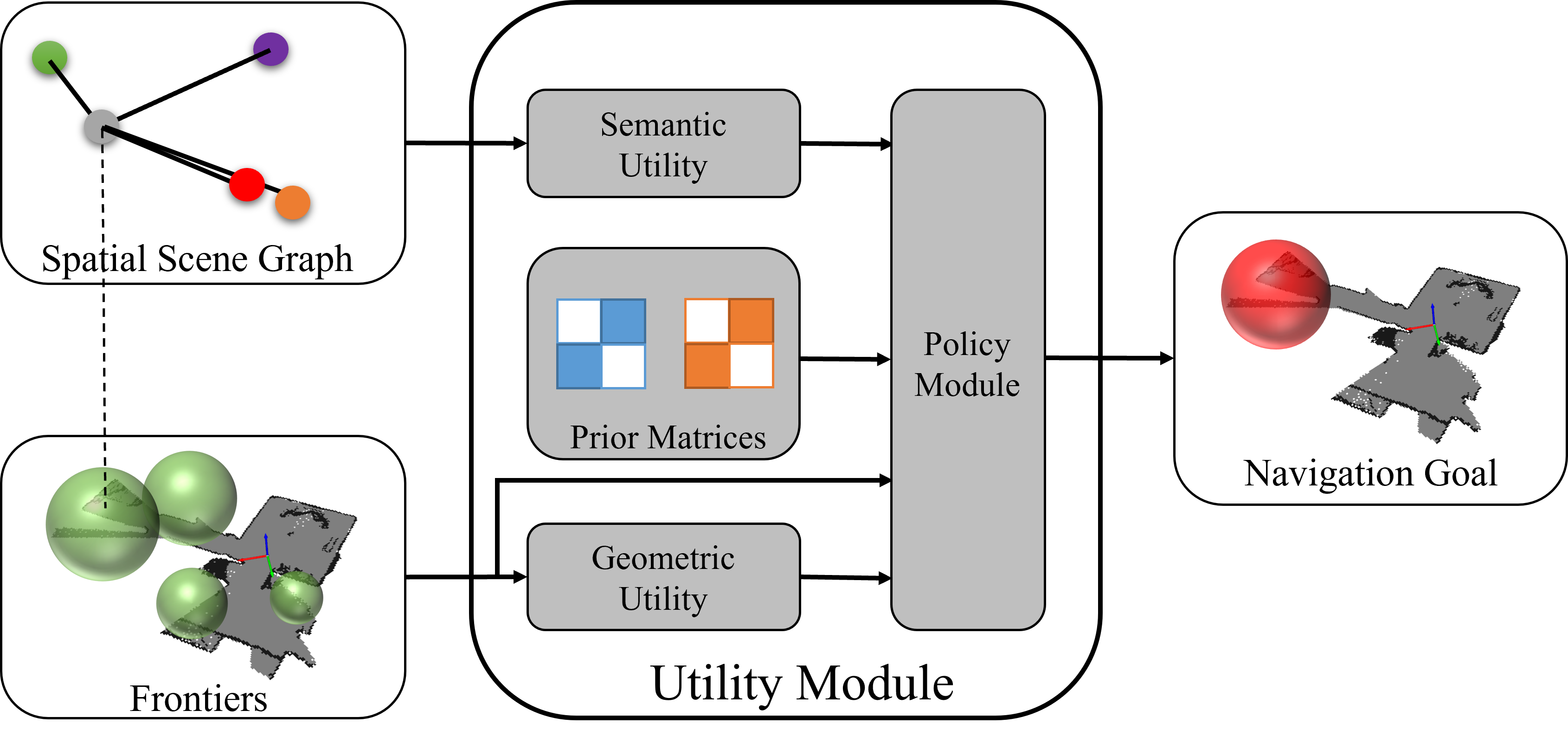}
         \caption{Utility Module}
         \label{fig:utility_model}
     \end{subfigure}
     \begin{subfigure}[b]{0.48\textwidth}
         \centering
         \includegraphics[width=\textwidth]{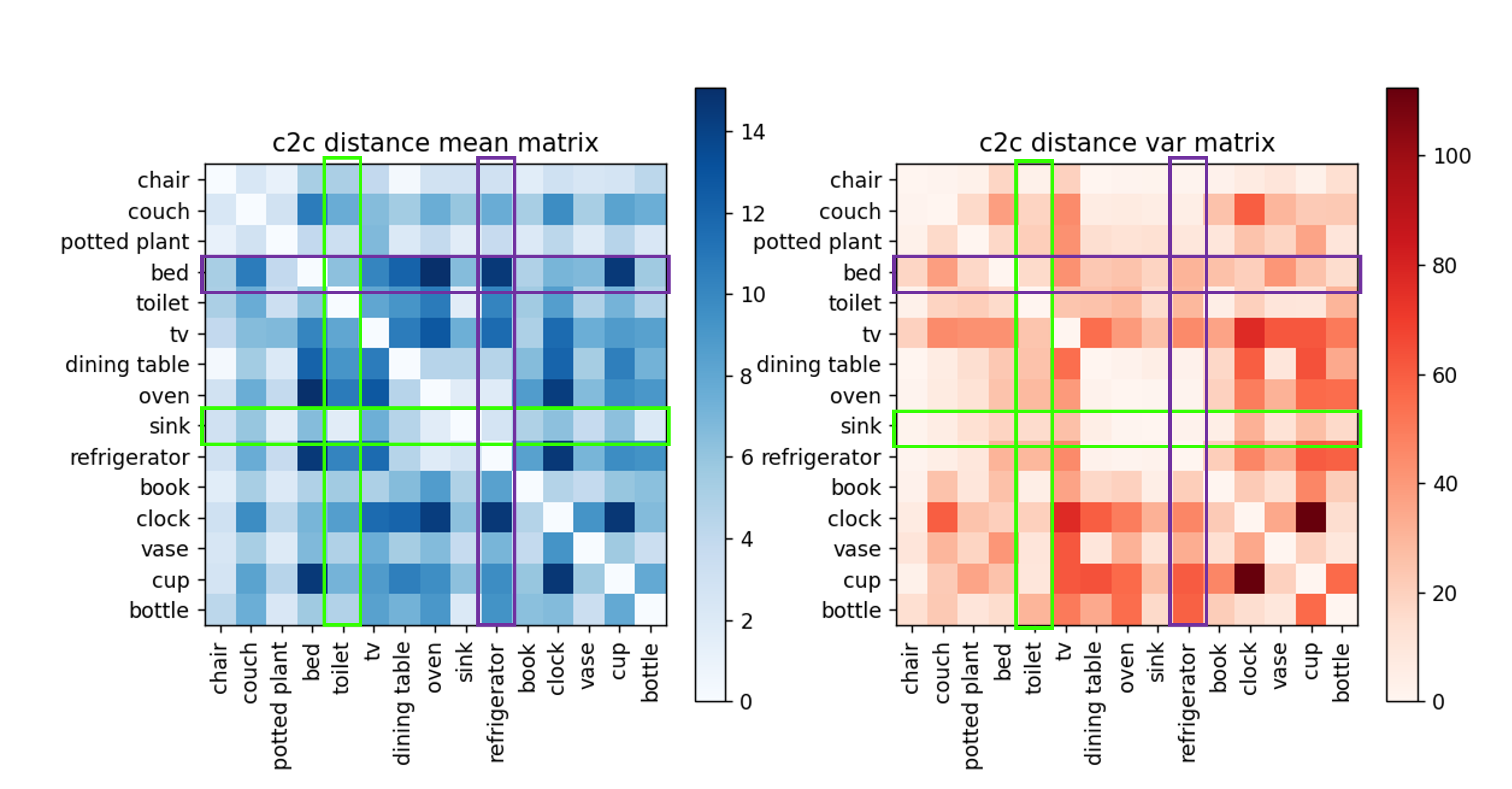}
         \caption{Prior Matrices}
         \label{fig:prior_matrices}
     \end{subfigure}
    \caption{\small \textbf{Utility Module and Prior Matrices.} a) Our utility module calculates the semantic utility from the spatial scene graph and the geometric utility from the frontiers, respectively. A policy module calculates the most promising frontier as the temporary navigation goal, based on the utilities and prior matrices. b) Prior matrices comprise a category-to-category prior distance matrix $\mD_{prior}$ and a category-to-category prior distance variance matrix $\mV_{prior}$. The green cross highlights the relationship between \textit{sink} and \textit{toilet}, that they have close proximity with high confidence. The purple cross highlights the relationship between \textit{bed} and \textit{refridgerator}, that they are always far from each other. }
    \label{fig.3}
\end{figure*}

\figLabel \ref{fig:utility_model} demonstrates how our utility module generates an intermediate navigation goal $\vx_t^{\text{goal}}$ in the exploration stage. Inspired by the idea of frontier-based exploration \cite{yamauchi97frontier}, our pipeline generates the frontiers $\mF_t=\{\vf_t^0, \cdots, \vf_t^{N_t}\}$ from the 2D occupancy map $\mM_t$ as intermediate navigation goal candidates. Each frontier $\vf_t^i=[x_t^i,y_t^i,l_t^i]\in\Re^3$ is composed of its position $[x_t^i,y_t^i]$ and frontier length $l_t^i$. At the beginning of exploration, the limited observations could provide very little semantic information about the scene so that the agent will explore the environment with the geometric utility, that is, to take the frontier with the maximum utility as the navigation goal. The geometric utility of frontier $\vf_t^i$ is defined as 

\begin{align}
u_{t, geo}^i=\frac{l_t^i}{dist(\hat{\bm{s}}_t^l, \vf_t^i, \mM_t)}
\end{align}
where $dist(\hat{\bm{s}}_t^l, \vf_t^i, \mM_t)$ is the geodesic distance from the current agent state to frontier $\vf_t^i$ on the 2D occupancy map $\mM_t$. This heuristic function, which derives a larger value with a larger frontier size and a shorter distance, describes the score of a frontier in greedy exploration policy. 

However, since our task is to navigate to the target object with the shortest path, instead of exploring as much area of the scene as possible, we propose to use novel utility functions to exploit the semantic information in the explored partial scene, to avoid unnecessary exploration. For this purpose, We calculate the semantic utility for each frontier by propagating semantic information of object nodes near the frontier to the frontier as its semantic utility. Specifically, we try to leverage the spatial relations between the observed objects in the scene with the target object category. For example, a frontier close to a basin and a toilet should be promising for finding a bathtub. The semantic utility of frontier $\vf_t^i$ is defined as 
\begin{align}
u_{t, sem}^i=\frac{\vr_t^i\cdot\vw_t^i}{k\cdot dist(\hat{\bm{s}}_t^L, \vf_t^i, \mM_t)}
\end{align}

where $\vr_t^i\in\Re^k$ is the relation scores vector of the $k$ objects around frontier $\vf_t^i$. The relation score for object category $j$ to the target category $c$ is $r_{jc}=1/d_{jc}$, that is, the inverse of the prior distance between category $j$ and $c$. Similarly, $\vw_t^i\in \Re^k$ is the discount weights vector of the $k$ objects, and the weight of object category $j$ with target category $c$ is $w^{jc}=1/sqrt(v_{jc})$. $\mD_{prior}=\{d_{jc}\}\in\Re^{N_c\times N_c}$ is the pre-computed category-to-category prior distance matrix. $\mV_{prior}=\{v_{jc}\}\in\Re^{N_c\times N_c}$ is the pre-computed category-to-category distance variance matrix, which serves to reduce the weight of the objects having little spatial relations to the target category. \figLabel \ref{fig:prior_matrices} also intuitively demonstrates how the pre-computed prior matrices help in general. The green cross highlights the relationship between \textit{sink} and \textit{toilet}, that they have a relatively small average distance $d_{jc}$ and a relatively small variance $v_{jc}$. This indicates the prior that \textit{sink} and \textit{toilet} are close with high confidence. For the same reason, the purple cross highlights the relationship between \textit{bed} and \textit{refrigerator}, that they are far away from each other with high probability, since they have a large prior distance $d_{jc}$ and small variance $v_{jc}$.


For the prior matrices used by the utility module, we collect data from three different sources \ie, BERT \cite{kenton2019bert}, CLIP \cite{radford2021clip} and Gibson \cite{Xia2018GibsonER}. With BERT and CLIP, we use pre-trained models to embed the class name strings and calculate the inter-class distance in the word embedding space as the prior distance. With Gibson, we calculate inter-class distance \textit{only on the train scenes} as the prior distance. Besides, we also calculate the inter-class distance variance matrix on Gibson, which is used to weigh the prior distances.


When it comes to the question of which objects to propagate semantic information to a frontier, we investigate two methods: 1) only select objects within a radius to a frontier and 2) use all objects while adding the inverse distance from an object to the frontier center as an extra weight factor to the semantic utility (soft radius method). For the policy module, we simply have the agent navigate by the geometric utility for the first 50 steps to collect enough observations, and then the agent navigates by semantic utility purely until the end of the episode.

\label{sec:navigation}

\section{Experiments}

\subsection{Experiment Setup}
\label{exp:setup}
We use Gibson \cite{Xia2018GibsonER} dataset in AI Habitat \cite{Savva2019HabitatAP} simulator in our experiments. Specifically, we follow the same settings as SemExp \cite{chaplot2020object}, using Gibson tiny split with 25 training scenes and 5 test scenes. Each test scene has 200 episodes. Each episode contains the starting point and the goal category. The episodes are provided by SemExp \cite{chaplot2020object}. The Habitat simulator APIs compute the ground truth path from a starting point to the closest goal category. For the perception setting, we use the simulated RGBD camera with resolution 640x480, field of view \ang{79}, and max depth range of 5 meters. We run all experiments on a Ubuntu 20.04 workstation with a single Nvidia RTX3080 graphics card. 

In our experiments, we first evaluate our pipeline under the same setting with SemExp \cite{chaplot2020object} to demonstrate that our method achieves state-of-the-art performance on the task of ObjectNav, even without performing any training on the given 25 training scenes. Then we present a comprehensive ablation study in two aspects. 1) Firstly, We conduct ablation studies on our semantic utility method to demonstrate how semantic information and other components help in this task. 2) Secondly, to better understand the problem essence and spot the bottleneck of ObjectNav, we present a comprehensive analysis of our modular pipeline: we provide ground truth data to different components of our pipeline and analyze how error accumulates through the data flow.

In the results, we report the results on the metrics proposed by Anderson \etal \cite{Anderson2018OnEvalEmbodiedAI} and also used by Habitat Challenge \cite{batra2020objectnav}, the \textit{Success Rate}, \textit{Success weighted by Path Length} (SPL) and \textit{Distance to Goal} (DTG). The \textit{Success Rate} is the ratio of episodes that the agent successfully navigates to the goal among all test episodes. SPL is defined as 
\begin{align}
    &\mathrm{SPL}=\frac{1}{N} \sum_{i=1}^{N} S_{i} \cdot \frac{\ell_{i}}{\max \left(p_{i}, \ell_{i}\right)}
\end{align}
where, $l_i$ is the length of the shortest path between the goal and the target for an episode, $p_i$ is the length of the path taken by the agent in an episode, and $S_i$ is the binary success indicator of the episode. DTG is the average agent's distance to the target object at the end of the episode.

\subsection{Implementation Details}
\label{exp:implement}


We use RTAB-Map \cite{Labbe2019RTABMap} to predict the agent pose and generate a 3D semantic reconstruction of the scene and a 2D occupancy map. When external ground truth odometry is provided, only the map assembler and 
 the map optimizer nodes in RTAB-Map, which are used for point cloud registration and global optimization, are used. Since RTAB-Map does not have native support for semantic SLAM, we launch another standalone RTAB-Map instance to reconstruct the semantic point cloud by 3D fusion. 
 


For the scene graph extraction module, we use DBSCAN \cite{ester1996dbscan} to cluster the semantic point cloud to object instances. The maximum distance between two points to be considered neighbors is $1.0$, and the minimum number of points in a neighborhood for a point to be considered a core point is $5$. All clusters with less than $5$ points will be filtered out as outliers. Then we further reduce the false detected instances by non-maximum suppression on all 3D bounding boxes with a threshold of $0.4$.


\subsection{Object Goal Navigation on Gibson}
\label{exp: object goal navigation}
In this section, at each step, all agents are given the ground truth pose and share the same Mask R-CNN \cite{he2017mask} model with pre-trained checkpoints provided by Detectron2 \cite{wu2019detectron2} to predict the corresponding semantic image from the RGB input, and share the same local planner to generate an instant action for the current step, to make a fair comparison. We report our experimental results along with two other recent works, as shown in table \ref{tab.1}. For SemExp\cite{chaplot2020object}, we report their improved results updated in Github. For RegQLearn \cite{gireesh22objQlearn}, we report the numbers in their paper.

\begin{table} [h]
\caption{\textbf{Results and Method Ablations.} Evaluated on 1000 episodes in Gibson validation split. Numbers reported on three metrics: success rate (Success), success-weighted path length (SPL), distance to goal in meters (DTG)}
\centering
\resizebox{0.9\columnwidth}{!}{
\begin{tabular}{@{}l l l l }
\toprule%
\bfseries Method & \bfseries Success & \bfseries SPL & \bfseries DTG \\

\midrule

SemExp \cite{chaplot2020object} & 0.650 & 0.330 & 1.576 \\
RegQLearn \cite{gireesh22objQlearn} & 0.637 & 0.313 & 1.568 \\
\myrowcolor%
Ours (BERT SemUtil) & \highest{0.693} & \highest{0.405} & 1.488\\

Ours (Gibson SemUtil) &\highest{0.693} & 0.383 & \highest{1.468} \\ 
\hhline{====}
\bfseries Ablations \\

(GeoUtil) & 0.623 &	0.359 &	1.890 \\

(CLIP SemUtil) & 0.677 & 0.375 & 1.606 \\

(SemUtil w/o.Var) & 0.689 & 0.383 & 1.528 \\

(SemUtil w/o.3DNMS) & 0.684	& 0.379 & 1.519\\

(SemUtil w/o.SoftR) & 0.624 & 0.344 & 1.839 \\

\bottomrule
\end{tabular}
}
\label{tab.1}
\end{table}

As demonstrated in the table, our method (StructNav) achieves state-of-the-art performance on all three metrics 
 with a fundamental margin on \textit{Success Rate} and \textit{Success weighted by Path Length} (SPL). Our method outperforms the best previous method SemExp by $6.6\%$ relatively on the success rate metric, by $22.7\%$ relatively on the SPL metric, and by $6.9\%$ relatively on the DTG metric.   

\subsection{Ablation Study on Semantic Utility Method}
\label{exp: method ablation}
In this section, we present the comprehensive ablations study on the utility modules of semantic frontiers. We provide the following ablations: \textit{\textbf{(i)}} GeoUtil: only use geometric utility for exploration, \ie, classical frontier-based exploration \cite{yamauchi97frontier}; \textit{\textbf{(ii)}} CLIP SemUtil: Use Language prior computed from CLIP \cite{radford2021clip}; \textit{\textbf{(iii)}} SemUtil w/o. Var: remove variance discount from BERT SemUtil; \textit{\textbf{(iv)}} SemUtil w/o. 3DNMS: remove 3D non-maximum suppression from BERT SemUtil and \textit{\textbf{(v)}} SemUtil w/o. SoftR: instead of using the soft radius method, compute the semantic utility of a frontier on objects within a radius to it with equal weights.
 
The result in \tblLabel \ref{tab.1} shows how different components in the utility module contribute to the final performance. Interestingly, our baseline of pure geometric-based frontier exploration achieves a very competitive result on Gibson. Besides, we observe a clear performance drop on (CLIP SemUtil) compared to (BERT SemUtil), which is also interesting since CLIP is trained by both texts and corresponding images. This could imply that the similarity in image features between categories does not necessarily relate to the spatial proximity in indoor scenes. 

\subsection{Qualitative Result of Semantic Utility Method}
\begin{figure*}[ht]
    \centering
     \begin{subfigure}[b]{0.46\textwidth}
         \centering
         \includegraphics[width=\textwidth]{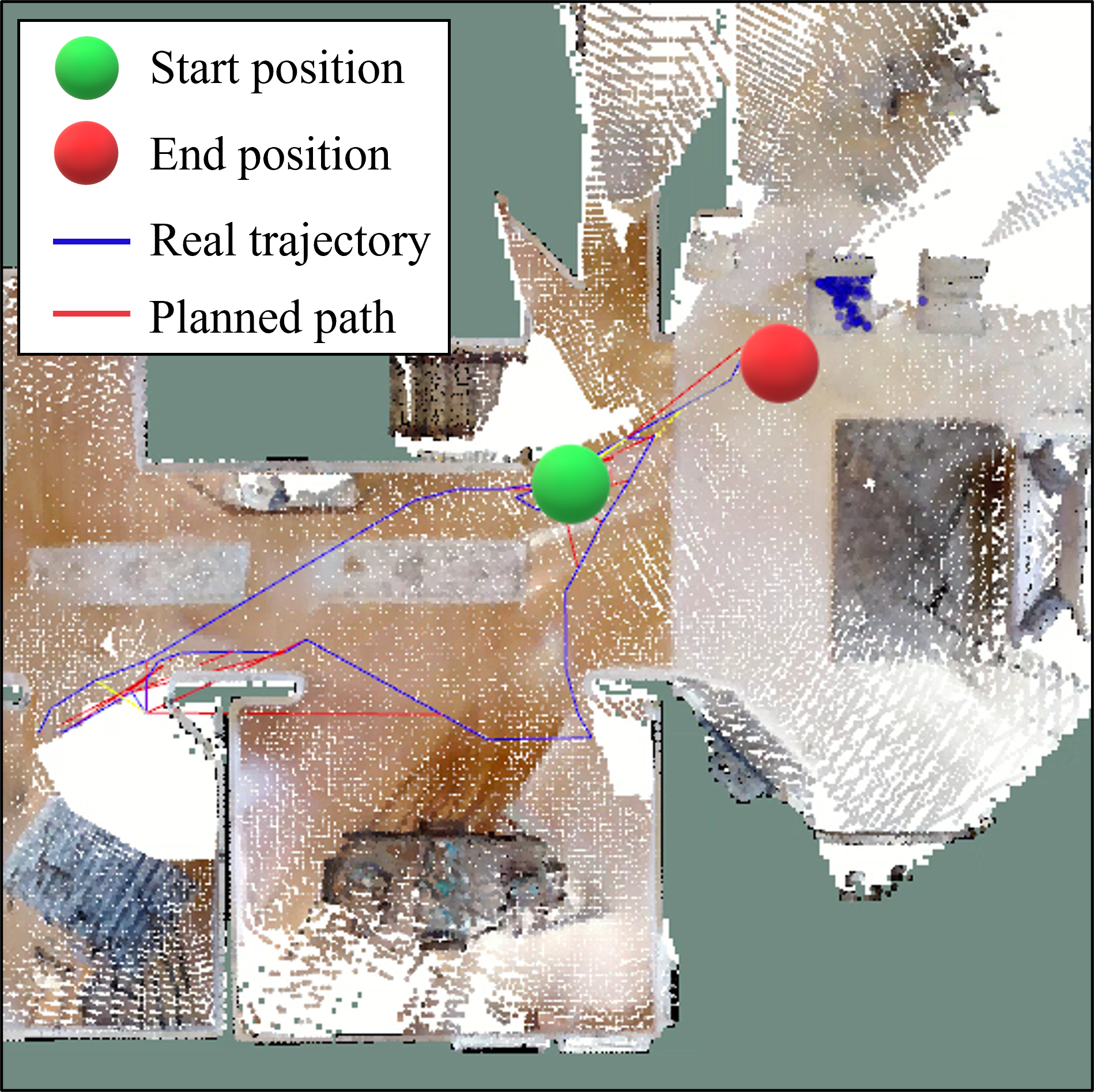}
         \caption{GeoUtil; Frontier-based Exploration\cite{yamauchi97frontier}}
         \label{fig:visualize_fexp}
     \end{subfigure}
     \begin{subfigure}[b]{0.46\textwidth}
         \centering
         \includegraphics[width=\textwidth]{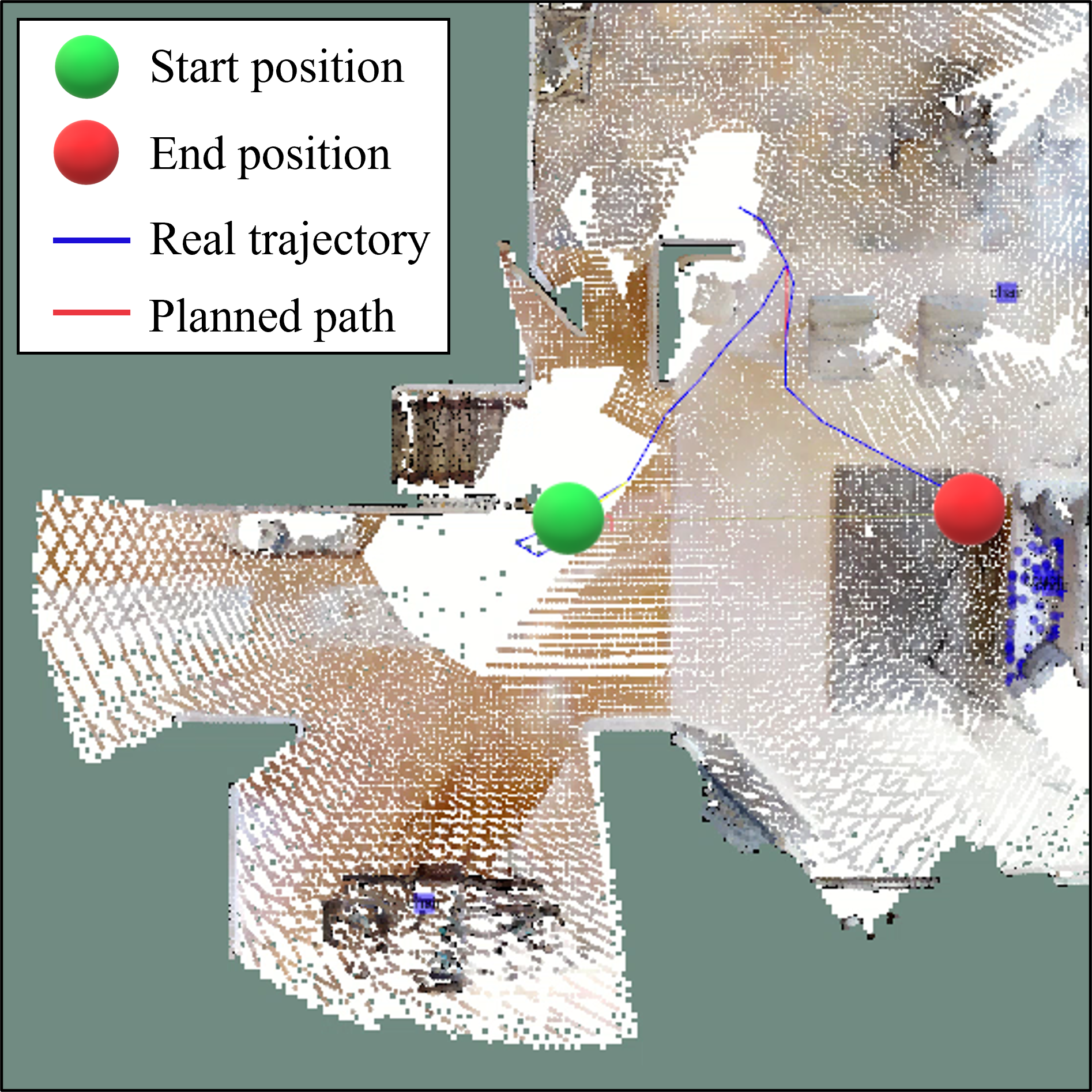}
         \caption{SemUtil; Our Method}
         \label{fig:visualize_sgnav}
     \end{subfigure}
    \caption{\textbf{Visualization of navigation trajectories in Rviz.}  A bird-eye view of the maps and trajectories on a test scene recovered using our approach. For this example, the result is obtained by running GeoUtil and SemUtil on the same episode to navigate to \textit{couch} on the test scene Wiconisco. The green sphere indicates the start of the episode. The red sphere suggests the end of this episode where the agent returns \textit{STOP} action to the simulator. The map's blue dots indicate this episode's detected target object. The blue lines are the real trajectory recorded from the TF frame attached to the agent base, while the red lines are the planned path from the planner.}\label{fig.4}
\end{figure*}

To further explain why and how semantic reasoning is useful in object goal navigation, we visualize and compare the trajectories of the navigation process with and without semantic utility, i.e. GeoUtil and SemUtil as defined in sections \ref{exp: object goal navigation}\ref{exp: method ablation}. The \figLabel \ref{fig.4} shows the visualization result of an episode to find \textit{couch} in the test scene Wiconisco. 

Compared with the trajectory from frontier-based exploration \ref{fig:visualize_fexp}, which is designed to explore as much as possible with the shortest path, we could see a clear shortcut from the result of our method \ref{fig:visualize_sgnav}. In fact, at the beginning of the episode, both agents rotate themselves to get observations from their surroundings. Then our method tries to follow the clue from semantic reasoning, navigating to the frontier most related to the target object, to avoid unnecessary exploration in the unknown environment. This qualitative result explains why our method gains a fantastic boost on the metric of \textit{Success weighted by Path Length} (SPL). This could also indicate that semantic reasoning and knowledge transferring are helpful for object goal navigation, and probably other indoor robotic tasks involving common sense knowledge. 

%


\subsection{Ablation Study on StructNav Pipeline}
\label{exp: pipeline ablation}
ObjectNav is an intricate problem that involves perception, understanding and interacting with the complex 3D physical world. It remains unclear what is the key bottleneck to solving this problem.
In this section, we present the ablation studies on the pipeline scale, i.e., how noise and errors in each module undermine the overall performance of the pipeline in the task of ObjectNav. 
Our modular pipeline allows us to do comprehensive ablation studies on each module. We inject the stepped ground truth data into the data flow of the pipeline (\figLabel \ref{fig.2}) and analyze how the performance improves if the pipeline uses ground truth data instead of predicted data from a module. The result is shown in \tblLabel \ref{tab.2}. In the table, \textbf{Odom.}, \textbf{SemSeg.}, \textbf{SG.} columns indicate the following experimental setting:

\begin{itemize}
    \item \textbf{Odom.}: GT. indicates using the ground truth pose of the agent; Pred. indicates using the predicted pose by RTAB-Map\cite{Labbe2019RTABMap}.
    \item \textbf{SemSeg.}: GT. indicates using the ground truth semantic segmentation provided by 3DSceneGraph\cite{armeni20193dscenegraph} dataset; Pred. indicates using the predicted semantic segmentation by Mask R-CNN\cite{he2017mask}. 
    \item \textbf{SG.}: GT. indicates using the ground truth scene graph from the Habitat simulator; Pred. indicates using the constructed scene graph by our method. For a fair comparison, objects in the unexplored area will be masked. Otherwise, the agent knows the ground-truth object's position in the unexplored area, which makes it close to the oracle.
\end{itemize}

It is also worth mentioning that our methods in \tblLabel \ref{tab.2} are directly comparable to the oracle. This is because the ground truth path is calculated by the Fast Marching Method (FMM) with the ground truth 2D occupancy map and ground truth start and end points. Thus, the methods presented in the \tblLabel \ref{tab.2} share the same global planner.

\begin{table}[h]
\caption{\textbf{Pipeline Ablations}. Results on ObjectNav with stepped ground truth data provided to the pipeline.}

\centering
\begin{tabular}{@{}l l l l l l l }
\toprule%
\bfseries Method & \bfseries Odom. & \bfseries SemSeg. &\bfseries SG. & \bfseries Success & \bfseries SPL & \bfseries DTG \\

\midrule
StructNav & Pred. & Pred. & Pred. & 0.576 & 0.340 & 2.123\\
\myrowcolor%
StructNav & GT. & Pred. & Pred. & 0.693 & 0.405 & 1.488 \\
\myrowcolor%
StructNav & GT. & GT. & Pred. & 0.842 & 0.563 & 1.098 \\
StructNav & GT. & GT. & GT. & 0.849 & 0.563 & 0.947 \\
Oracle & \textbackslash & \textbackslash & \textbackslash & 1.000 & 1.000 & 0.000\\
\bottomrule
\end{tabular}
\label{tab.2}
\end{table}


As highlighted in \tblLabel \ref{tab.2}, among the three ablated modules, the major bottleneck of StructNav is the semantic segmentation module which is a pre-trained MaskRCNN \cite{he2017mask} in our experiments. With a perfect semantic segmentation model, the success rate boosts about 15\% from 0.693 to 0.842, which indicates that a better semantic segmentation method is crucial to tackling the ObjectNav problem. Moreover, we conduct an ablation study on semantic segmentation using degraded models, elucidating the adverse impact of poor results in semantic segmentation on the overall system performance. Given spatial constraints, the results and discussions have been relocated to \appendixautorefname \ref{sec:ablation_semseg}.

The impact of the visual odometry module is also comparably large. The ground truth odometry lifts the success rate by $11.7\%$ from 0.576 to 0.693 and SPL by $0.065$ from 0.340 to 0.405. It is worth mentioning that although our experiment sheds some light on how visual odometry might impact the overall performance of object goal navigation to investigate the sim-to-real gap problem, there is still a gap between the experiment setting and the real robotic system. The simulation's action space is discrete, and it allows sliding when an agent moves against the boundary of the traversable space with a certain heading angle range. This gives more tolerance to localization errors in the simulation than in the real robotic system.

However, the scene graph extraction module shows little impact on the result, which indicates that with a perfect semantic segmentation model, the extracted spatial scene graph is "good enough" for navigation. The second last row shows the result with ground truth odometry, semantic segmentation, and scene graph, which still has a gap of $15.1\%$ in success rate and $0.437$ in SPL compared to the Oracle. The gap in success rate could be attributed to the errors in the SLAM system, especially the errors in the 2D occupancy map projected by the 3D reconstruction. The gap in SPL could be more attributed to the variance in scene layout. 






\section{Conclusion}


This paper proposes a training-free pipeline \textbf{StructNav} to solve the object goal navigation problem. Built on top of a visual SLAM system, our introduced pipeline constructs a structured representation of the scene, then navigates the agent to the target object by leveraging the information in the observed partial scene and the prior knowledge of the category-to-category spatial relations. Extensive experiments demonstrate that our training-free pipeline can reach state-of-the-art performance on the Gibson dataset with a large margin compared to previous training-heavy methods. Yet, from the pipeline ablations, we find that the semantic model could be a major bottleneck of this task. Therefore, it would be interesting to explore the semantic segmentation model and semantic frontier module as a future extension of our work.


\section{APPENDIX}
\label{sec:appendix}

\subsection{Ablation Study on Semantic Segmentation Models}
\label{sec:ablation_semseg}
\begin{figure}[ht]
    \centering
    \includegraphics[width=1\linewidth]{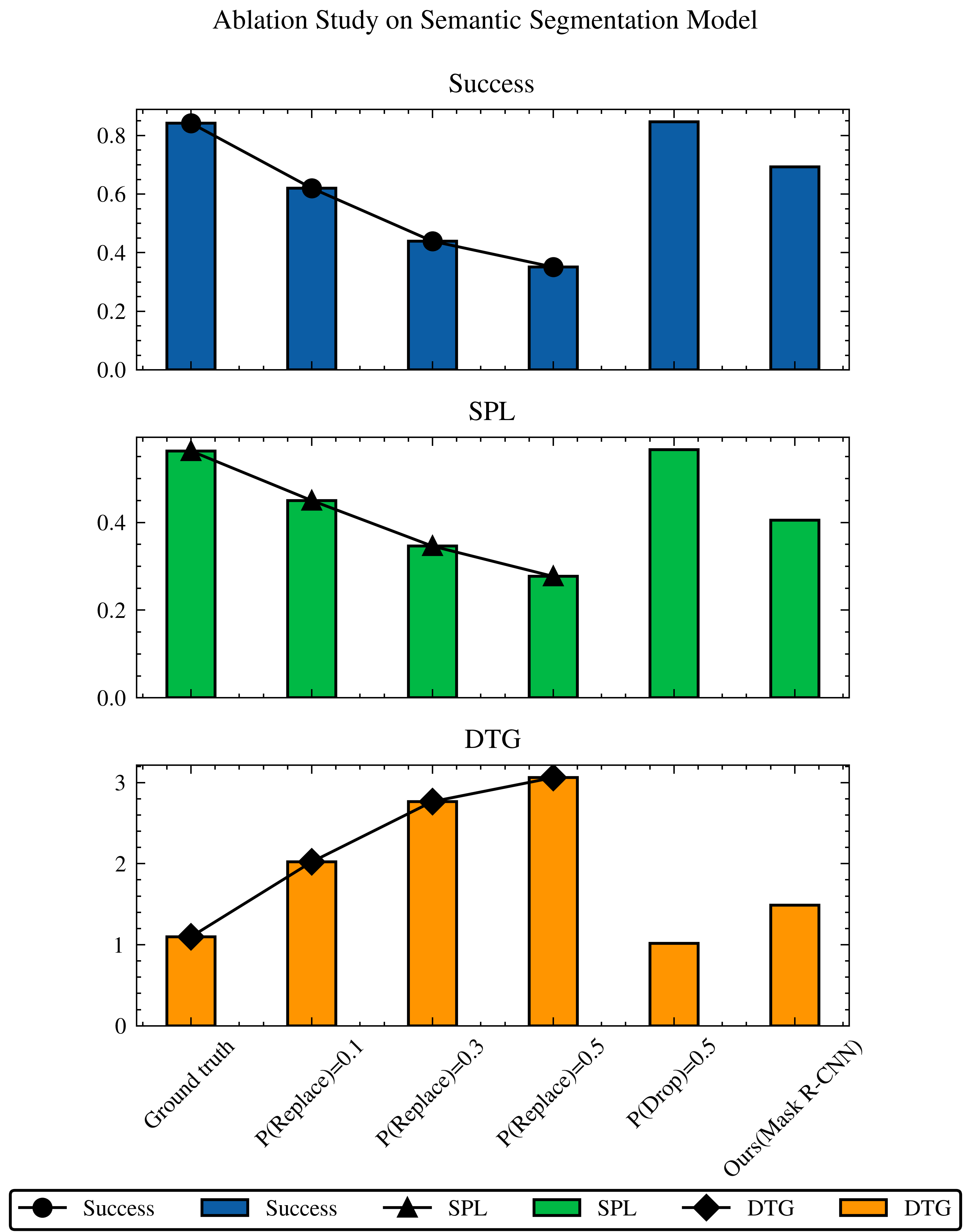}
    \vspace{1pt}
    \caption{\textbf{Ablation study on Semantic Segmentation Models.} The notation P(Replace)=X denotes the random replacement of the class label of an instance mask with another class label, where the replacement occurs with a probability of X. Similarly, P(Drop)=0.5 signifies the random dropping of the class label of an instance mask to the background, with a probability of 0.5. The line chart visually demonstrates the detrimental effects of heightened noise in semantic models on the system's performance across three key metrics: Success (Success Rate), SPL (Success Weighted by Path Length), and DTG (Distance to Goal). The bar charts offer an intuitive comparison of various semantic segmentation models. It is noteworthy that, in the case of Success and SPL, larger values indicate superior performance, while for DTG, higher values reflect worse performance.}
\label{fig.5}
\end{figure}

We conducted an ablation study to explore the impact of the semantic segmentation model on the general performance of our object goal navigation system, significantly how a degraded model undermines the system performance. For this purpose, we add artificial noise of different scales to the ground truth semantic segmentation model and report the system performance on the object goal navigation task. Specifically, ablations include \textit{\textbf{(i)}} Ground truth: Using the ground truth segmentation from simulation, which is identical to the third row reported in pipeline ablation \tblLabel \ref{tab.2}; \textit{\textbf{(ii)}} P(Replace)=0.1: Randomly replace the class label of an instance mask with another class label with probability 0.1; \textit{\textbf{(iii)}} P(Replace)=0.3: Randomly replace the class label of an instance mask with another class label with probability 0.3; \textit{\textbf{(iv)}} P(Replace)=0.5: Randomly replace the class label of an instance mask with another class label with probability 0.5; \textit{\textbf{(v)}} P(Drop)=0.5: Randomly drop the class label of an instance mask to the background with probability 0.5. This is also equivalent to reducing segmentation recall to 0.5 while keeping precision to 1.0; \textit{\textbf{(vi)}} Ours(Mask R-CNN): For comparison with real semantic segmentation model, we add the result with Mask R-CNN here. This line is identical to the second row reported in pipeline ablation \tblLabel \ref{tab.2}. The numeric results are in table \tblLabel \ref{tab.3}, and visualized in \figLabel \ref{fig.5}. 

\begin{table}[h]
\caption{\textbf{Ablation study on Semantic Segmentaion Models}. System performance with different degraded semantic segmentation models. The notation P(Replace)=X denotes the random replacement of the class label of an instance mask with another class label, where the replacement occurs with a probability of X. Similarly, P(Drop)=0.5 signifies the random dropping of the class label of an instance mask to the background, with a probability of 0.5. }

\centering
\resizebox{0.9\columnwidth}{!}{
\begin{tabular}{@{}l l l l l l l }
\toprule%
\bfseries Segmentation Model & \bfseries Success & \bfseries SPL & \bfseries DTG \\

\midrule
Ground truth & 0.842 & 0.563 & 1.098 \\
P(Replace)=0.1 & 0.620 & 0.450 & 2.022 \\ 
P(Replace)=0.3 & 0.439 & 0.346 & 2.766 \\
P(Replace)=0.5 & 0.351 & 0.277 & 3.061 \\
P(Drop)=0.5 & 0.847 & 0.566 & 1.017 \\
Ours(Mask R-CNN) & 0.693 & 0.405 & 1.488 \\
\bottomrule
\end{tabular}
}
\label{tab.3}
\end{table}

Our results showed that the performance of semantic segmentation models significantly impacts the object goal navigation system. A semantic segmentation model with an error rate of $50\%$ has less than $50\%$ performance of the system with ground truth segmentation model on all three metrics. An interesting finding is that randomly dropping segmentation labels has much less impact on the system performance. The explanation could be that since labels are randomly dropped, an object's probability of having no label in $N$ frames in sequence could be small. For example, let $N=5$, $P(\textit{no label in N frames})=0.5^5\approx0.031$. Thus, since we integrate semantic segmentation labels in the SLAM pipeline and use labels of point cloud for navigation, the impact is mitigated through time. Overall, our ablation study highlights the importance of semantic segmentation models in object goal navigation systems and further support our idea that semantic segmentation could be a major bottleneck for object goal navigation. 




\bibliographystyle{plainnat}
\bibliography{IEEEabrv, ref}


\end{document}